\documentclass[runningheads]{llncs}

\usepackage{iciap}

\usepackage{iciapabbrv}

\usepackage{graphicx}
\usepackage{booktabs}
\usepackage{multicol}
\usepackage{multirow}
\usepackage{caption}
\usepackage{comment}
\usepackage{adjustbox}
\usepackage[inkscape]{svg}
\usepackage{caption}
\usepackage{float}

\usepackage[accsupp]{axessibility}  

\usepackage{hyperref}

\usepackage{orcidlink}

\usepackage[dvipsnames]{xcolor}

\begin{document}

\newcommand{\methname}{ExDD\xspace}

\title{ExDD: Explicit Dual Distribution Learning for Surface Defect Detection via Diffusion Synthesis}
\titlerunning{ExDD: Explicit Dual Distribution Learning}

\author{Muhammad Aqeel\inst{1}\orcidlink{0009-0000-5095-605X} \and
Federico Leonardi\inst{1} \and
Francesco Setti\inst{1,2}\orcidlink{0000-0002-0015-5534}} 

\authorrunning{M. Aqeel et al.}

\institute{Dept. of Engineering for Innovation Medicine, University of Verona\\ 
Strada le Grazie 15, Verona, Italy\\
\and
Qualyco S.r.l., Strada le Grazie 15, Verona, Italy \\
Contact author: \email{muhammad.aqeel@univr.it}
}

\maketitle

\begin{abstract}

Industrial defect detection systems face critical limitations when confined to one-class anomaly detection paradigms, which assume uniform outlier distributions and struggle with data scarcity in real-world manufacturing environments. We present \methname (\textbf{Ex}plicit \textbf{D}ual \textbf{D}istribution), a novel framework that transcends these limitations by explicitly modeling dual feature distributions. Our approach leverages parallel memory banks that capture the distinct statistical properties of both normality and anomalous patterns, addressing the fundamental flaw of uniform outlier assumptions. To overcome data scarcity, we employ latent diffusion models with domain-specific textual conditioning, generating in-distribution synthetic defects that preserve industrial context. Our neighborhood-aware ratio scoring mechanism elegantly fuses complementary distance metrics, amplifying signals in regions exhibiting both deviation from normality and similarity to known defect patterns. Experimental validation on KSDD2 demonstrates superior performance (94.2\% I-AUROC, 97.7\% P-AUROC), with optimal augmentation at 100 synthetic samples. \url{https://github.com/aqeeelmirza/ExDD-Defect-Detection}
\end{abstract}

\section{Introduction}
\label{sec:intro}

Surface defect detection is a cornerstone of industrial quality control, where even microscopic imperfections in materials like copper, steel, or marble can lead to catastrophic failures in downstream applications \cite{aqeel2025CoMet, vrochidou2022marble, zhang2021visual}. Traditional computer vision approaches, 
struggle with the inherent variability of industrial defects, prompting a shift toward deep learning \cite{bhatt2021image}. However, supervised methods face a critical limitation: the scarcity of annotated defect data due to their rarity in production lines \cite{jawahar2023leather}. To address this, recent works like \cite{roth2022towards} and \cite{zavrtanik2021draem} have popularized one-class anomaly detection, which trains exclusively on normal samples. While effective in controlled settings, these methods implicitly assume anomalies are uniformly distributed outliers, a flawed premise for structured defects 
that occupy distinct distributions in feature space \cite{chen2021surface}.

The reliance on one-class paradigms also ignores a key insight: industrial defects often exhibit consistent patterns 
with distinctive visual characteristics that can be separated from normal textures when properly represented in feature space \cite{zavrtanik2022dsr}. Recent attempts to model anomaly distributions, such as patch-level density estimation~\cite{defard2021padim}, lack explicit distribution separation, while dual subspace re-projection~\cite{zavrtanik2022dsr} oversimplifies defect geometry. Furthermore, synthetic anomaly generation methods like \cite{jain2022synthetic} often produce out-of-distribution artifacts due to adversarial training or unrealistic perturbations, as noted by \cite{girella2024cbmi}. This misalignment between synthetic and real defects undermines feature learning, particularly for subtle anomalies 
\cite{capogrosso2024diffusion}.

Recent advances in diffusion models offer a promising solution. By leveraging text-conditional generation, where defect descriptions in natural language guide the synthesis process, \cite{girella2024cbmi} demonstrated that latent diffusion models (LDMs) can synthesize in-distribution defects that preserve the statistical properties of real anomalies. However, their framework treats synthesis as a preprocessing step, decoupling it from the detection pipeline. 
Meanwhile, self-supervised methods like \cite{aqeel2024delta} improve robustness through pretext tasks but fail to explicitly model the defect distribution, resulting in suboptimal separability.

In this paper, we propose $\methname$ (\textbf{Ex}plicit \textbf{D}ual \textbf{D}istribution), a unified framework that bridges explicit dual distribution modeling 
and diffusion-based defect synthesis for surface inspection. Unlike prior work, \methname jointly optimizes two memory banks: (1) a normal memory encoding nominal feature distributions and (2) a defect memory populated by diffusion-synthesized anomalies. The synthesis process uses text prompts derived from domain expertise (e.g., ``metallic scratches'') to generate defects that align with the true anomaly distribution, as validated by \cite{capogrosso2024diffusion}. Crucially, our dual memory architecture enables neighborhood-aware ratio scoring, which amplifies deviations from normality while suppressing false positives caused by normal feature variations a common failure mode in one-class methods \cite{defard2021padim}.

The main contributions of our paper are threefold:
\begin{itemize}
    \item \textbf{Dual Distribution Learning:} We formalize surface defect detection as a dual feature distribution separation problem, explicitly modeling both normal and defect feature distributions via memory banks.
    \item \textbf{Diffusion-Augmented Training:} A text-conditional LDM synthesizes in-distribution defects, expanding the defect memory while preserving geometric fidelity.
    \item \textbf{Ratio Scoring:} A novel scoring mechanism combines distance-to-normal and similarity-to-defect metrics, leveraging the dual memory structure for robust decision boundaries.
\end{itemize}

\section{Related Work}
\label{sec:relatedwork}

Surface defect detection has advanced through three interconnected research streams: one-class normality modeling, synthetic defect generation, and self-supervised feature learning. One-class anomaly detection dominates industrial applications due to data scarcity, with methods like PatchCore \cite{roth2022towards} using memory banks of nominal features and PaDiM \cite{defard2021padim} modeling patch-wise distributions. However, these approaches struggle with structured defects like scratches that occupy distinct feature distributions \cite{chen2021surface}, as highlighted by failures in detecting fine marble cracks \cite{vrochidou2022marble}.

Synthetic data generation addresses annotation scarcity but risks producing unrealistic artifacts. Early methods using random noise \cite{jain2022synthetic} often distort defect semantics, while GAN-based approaches improved realism but suffered from mode collapse \cite{akcay2019ganomaly}. Diffusion models offer a robust alternative, with \cite{capogrosso2024diffusion} generating defects via text prompts aligned with domain expertise. However, most methods decouple synthesis from detection, preventing joint optimization, a gap addressed by \methname's integrated framework.

Self-supervised methods learn features without defect labels through pseudo-label refinement \cite{aqeel2024delta} and meta-learning for threshold adaptation~\cite{aqeel2025CoMet, aqeel2024meta}, though many require partial annotations \cite{bovzivc2021mixed}. Hybrid approaches like DRAEM \cite{zavrtanik2021draem} train discriminative reconstructions but ignore defect feature structures. Recent work with latent diffusion models \cite{girella2024cbmi} synthesizes in-distribution defects but decouples synthesis from detection. In contrast, \methname unifies self-supervised principles with explicit modeling of separate normal and anomalous distributions, ensuring synthetic and real defects form a cohesive anomaly subspace while bridging generation and detection through an integrated framework.

\section{\methname Framework}
\label{sec:method}

We propose \methname, a dual memory bank paradigm that extends memory-based anomaly detection by explicitly modeling both normal and anomalous feature distributions. In this section, we formalize the problem setup (section~\ref{sec:problem}), describe our dual memory bank architecture (section~\ref{sec:dual_memory}), detail our diffusion-based synthetic anomaly generation approach (section~\ref{sec:diffusion}), and present our novel anomaly scoring mechanism (section~\ref{sec:anomaly_detection}).

\begin{figure}[tb]
  \centering
  \includegraphics[width=0.8\textwidth]{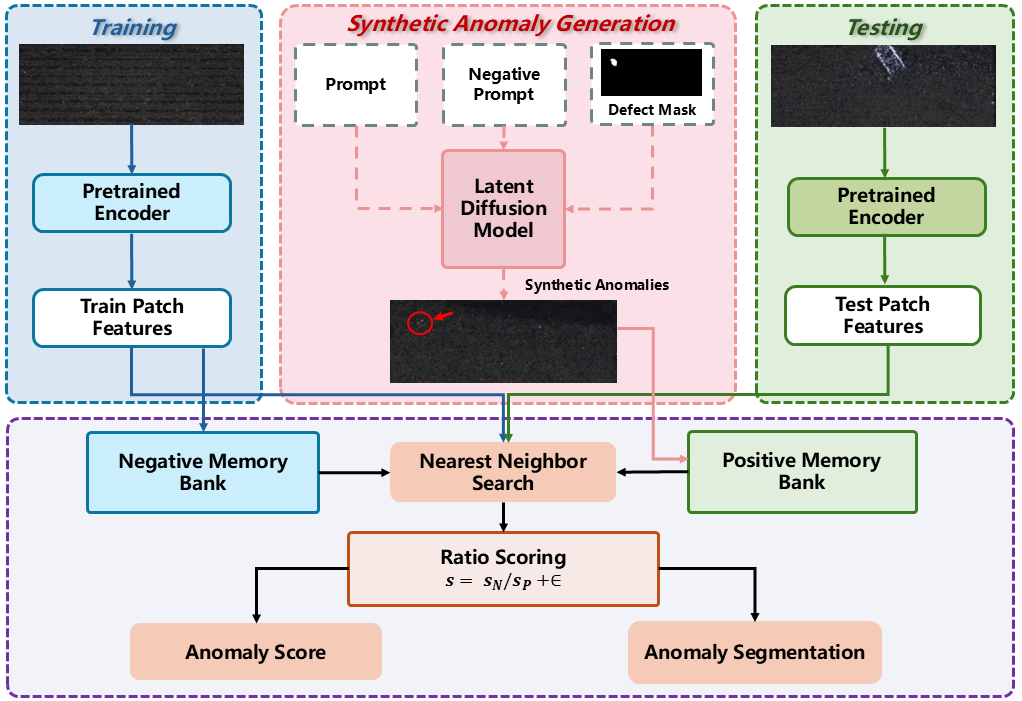}
  \caption{Overview of the \methname framework, illustrating the training process with pretrained encoder and patch feature extraction, synthetic anomaly generation using diffusion models with prompt guidance, testing workflow, and the dual memory bank architecture with ratio-based anomaly scoring mechanism.
  }
  \label{fig:method}
\end{figure}

\subsection{Problem Formulation}
\label{sec:problem}

Let $\mathcal{X}_N$ denote the set of nominal images ($\forall x \in \mathcal{X}_N : y_x = 0$) available during training, where $y_x \in \{0, 1\}$ indicates whether an image $x$ is nominal (0) or anomalous (1). Similarly, $\mathcal{X}_T$ represents the test set, with $\forall x \in \mathcal{X}_T : y_x \in \{0, 1\}$. Let $\mathcal{X}_A$ denote the set of anomalous samples, which may be available in limited quantity or generated synthetically.

Traditional anomaly detection operates in a one-class paradigm, modeling only the distribution of normality $P(\mathcal{X}_N)$ and measuring deviations from this established distribution. This approach assumes anomalies are uniformly distributed in the complement space, which is not true for industrial defects with consistent patterns. Our key insight is that industrial anomalies often form distinct distributions in feature space. By explicitly modeling both distributions, we create a more discriminative decision boundary.

Following established protocols [1,2], we use a network $\phi$ pre-trained on ImageNet as our feature extractor. We denote $\phi_{i,j} = \phi_j(x_i)$ as the features for image $x_i \in \mathcal{X}$ at hierarchy level $j$ of network $\phi$, where $j \in \{1,2,3,4\}$ typically indexes feature maps from ResNet architectures.

\subsection{Dual Memory Bank Architecture}
\label{sec:dual_memory}

The core innovation of \methname is its parallel memory bank architecture, which explicitly models normal and anomalous feature distributions.

\subsubsection{Locally Aware Patch Features}

To extract an informative description of patches, we employ the local patch descriptors defined in~\cite{roth2022towards}.
For a feature map tensor $\phi_{i,j} \in \mathbb{R}^{c^* \times h^* \times w^*}$ with depth $c^*$, height $h^*$, and width $w^*$, we denote $\phi_{i,j}(h, w) = \phi_j(x_i, h, w) \in \mathbb{R}^{c^*}$ as the $c^*$-dimensional feature slice at position $(h,w)$. To incorporate local spatial context, we define the neighborhood of position $(h,w)$ with patch size $p$ as:
\begin{equation}
\mathcal{N}_p^{(h,w)} = \{(a, b) | a \in [h - \lfloor p/2 \rfloor, ..., h + \lfloor p/2 \rfloor], b \in [w - \lfloor p/2 \rfloor, ..., w + \lfloor p/2 \rfloor]\}
\end{equation}

The locally aware patch features at position $(h,w)$ are computed as:
\begin{equation}
\phi_{i,j}(\mathcal{N}_p^{(h,w)}) = f_{\text{agg}}(\{\phi_{i,j}(a,b) | (a,b) \in \mathcal{N}_p^{(h,w)}\})
\end{equation}
where $f_{\text{agg}}$ is an aggregation function, implemented as adaptive average pooling with a $3 \times 3$ window size. For a feature map tensor $\phi_{i,j}$, its collection of locally aware patch features is:
\begin{equation}
\mathcal{P}_{s,p}(\phi_{i,j}) = \{\phi_{i,j}(\mathcal{N}_p^{(h,w)}) | h, w \mod s = 0, h < h^*, w < w^*, h, w \in \mathbb{N}\}
\end{equation}
where $s$ is a stride parameter (set to 1 in our implementation).

We extract features from both layer 2 and layer 3 of the backbone network. Features from layer 3 are upsampled to match layer 2 dimensions, then concatenated:
\begin{equation}
\mathcal{P}_{s,p}(\phi_{i,\{2,3\}}) = \text{Concat}(\mathcal{P}_{s,p}(\phi_{i,2}), \text{Upsample}(\mathcal{P}_{s,p}(\phi_{i,3})))
\end{equation}

\subsubsection{Negative and Positive Memory Banks}

Unlike traditional one-class methods, \methname leverages the statistical properties of both normal and anomalous features through parallel memory banks. The Negative Memory Bank ($\mathcal{M}_N$) stores patch-level features from nominal samples:
\begin{equation}
\mathcal{M}_N = \bigcup_{x_i \in \mathcal{X}_N} \mathcal{P}_{s,p}(\phi_{\{2,3\}}(x_i))
\end{equation}

Complementarily, the Positive Memory Bank ($\mathcal{M}_P$) stores patch-level features from anomalous samples:
\begin{equation}
\mathcal{M}_P = \bigcup_{x_i \in \mathcal{X}_A} \mathcal{P}_{s,p}(\phi_{\{2,3\}}(x_i))
\end{equation}

The deliberate separation of memory banks preserves the distinct statistical properties of normal and anomalous feature distributions. Please note that the positive memory bank only accounts for those patches related to the defects. While this would require a pixel-level annotation of all the defects in the case of real images, it comes for free in the case of synthetic images, where the localization of defective patches can be automated by simply computing the difference between the original and generated images.

\subsubsection{Dimensionality Reduction and Coreset Subsampling}

The concatenated feature vectors have a high dimensionality of 1536 channels. We apply random projection based on the Johnson-Lindenstrauss lemma:
\begin{equation}
\psi: \mathbb{R}^d \rightarrow \mathbb{R}^{d^*}
\end{equation}
where $d^* = 128 < d = 1536$. The projection matrix is constructed with elements drawn from a standard Gaussian distribution and normalized to unit length.

Even after dimensionality reduction, we employ greedy coreset subsampling:
\begin{equation}
\mathcal{M}_C^* = \arg\min_{\mathcal{M}_C \subset \mathcal{M}} \max_{m \in \mathcal{M}} \min_{n \in \mathcal{M}_C} \|m - n\|_2
\end{equation}

This objective ensures that the selected coreset provides optimal coverage of the feature space.

We apply asymmetric subsampling: 2\% for the negative memory bank (higher redundancy) and 10\% for the positive bank (preserve diverse anomaly representations).

\subsection{Diffusion-based Anomaly Generation}
\label{sec:diffusion}

To address the limited availability of anomalous samples, we implement a diffusion-based data augmentation pipeline inspired by DIAG~\cite{girella2024cbmi}.

\subsubsection{Theoretical Foundation}

DIAG leverages Latent Diffusion Models (LDMs) to generate synthetic anomalies in a lower-dimensional latent space. The data distribution $q(x_0)$ is modeled through a latent variable model $p_\theta(x_0)$:
\begin{equation}
p_\theta(x_0) = \int p_\theta(x_{0:T})dx_{1:T}
\end{equation}
\begin{equation}
p_\theta(x_{0:T}) := p_\theta(x_T)\prod_{t=1}^T p_\theta^{(t)}(x_{t-1}|x_t)
\end{equation}

The parameters $\theta$ are learned by maximizing an ELBO of the log evidence:
\begin{equation}
\max_\theta \mathbb{E}_{q(x_0)}[\log p_\theta(x_0)] \leq \max_\theta \mathbb{E}_{q(x_0,x_1,...,x_T)} [\log p_\theta(x_{0:T}) - \log q(x_{1:T}|x_0)]
\end{equation}
where $q(x_{1:T}|x_0)$ is a fixed inference process defined as a Markov chain.

By conditioning on normal images and anomaly masks derived from real defects, our generative process creates samples within the true distribution of industrial defects. This ensures the positive memory bank captures genuine anomaly patterns rather than artifacts.
\subsubsection{Synthetic Anomaly Generation Pipeline}

To generate an anomalous image $i_a$, we start with a triplet $(i_n, d_a, m_a)$ consisting of a nominal image $i_n \in \mathcal{X}_N$, a textual anomaly description $d_a$, and a binary mask $m_a$. We utilize Stable Diffusion XL's inpainting capabilities with prompts like ``copper metal scratches'' and ``white marks on the wall'' based on KSDD2 dataset analysis. The pipeline uses inference steps=30, guidance scale=20.0, strength=0.99, and padding mask crop=2, resulting in a robust positive memory bank capturing diverse anomaly patterns.

\subsection{Anomaly Detection with \methname}
\label{sec:anomaly_detection}

The key innovation in \methname's detection mechanism is its ability to measure both dissimilarity from normality and similarity to anomaly patterns.

\subsubsection{Distance Computation}

For a test image $x_{\text{test}}$, we compute two complementary distance measures:
\begin{enumerate}
\item \textbf{Negative Distance} $(s_N^*)$: The maximum minimum Euclidean distance from test patch features to the negative memory bank:
\begin{align}
m_{\text{test},*}, m^* &= \arg\max_{m_{\text{test}} \in \mathcal{P}(x_{\text{test},*})} \arg\min_{m \in \mathcal{M}_N} \|m_{\text{test}} - m\|_2 \\
s_N^* &= \|m_{\text{test},*} - m^*\|_2
\end{align}

\item \textbf{Positive Distance} $(s_P^*)$: The maximum minimum Euclidean distance to the positive memory bank:
\begin{align}
m_{\text{test},+}, m^+ &= \arg\max_{m_{\text{test}} \in \mathcal{P}(x_{\text{test}})} \arg\min_{m \in \mathcal{M}_P} \|m_{\text{test}} - m\|_2 \\
s_P^* &= \|m_{\text{test},+} - m^+\|_2
\end{align}
\end{enumerate}

\subsubsection{Neighborhood-Aware Weighting}

To account for the local neighborhood structure in feature space, we incorporate neighborhood-aware weighting based on local density estimation theory:

\begin{equation}
w_N^* = 1 - \frac{e^{-s_N^*/\sqrt{d}}}{\sum_{m \in \mathcal{N}_b(m^*)} e^{-\|m_{\text{test},*} - m\|_2/\sqrt{d}}}
\end{equation}

This formulation increases the weighting factor when a test patch's nearest neighbor in the normal memory bank is isolated from other normal features.

For the positive raw anomaly score, we invert the formulation:
\begin{equation}
w_P^* = \frac{e^{-s_P^*/\sqrt{d}}}{\sum_{m \in \mathcal{N}_b(m^+)} e^{-\|m_{\text{test},+} - m\|_2/\sqrt{d}}}
\end{equation}

The weighted scores are computed as:
\begin{align}
s_N &= w_N^* \cdot s_N^* \\
s_P &= w_P^* \cdot s_P^*
\end{align}
where $\mathcal{N}_b(m^*)$ and $\mathcal{N}_b(m^+)$ represent the $b$ nearest neighbors to $m^*$ and $m^+$ in their respective memory banks.

\subsubsection{Ratio Scoring}

We introduce a novel Ratio Scoring method that fuses information from both memory banks:
\begin{equation}
s_{\text{ratio}} = \frac{s_N}{s_P + \epsilon}
\end{equation}
where $\epsilon$ is an arbitrary small constant value added to prevent division by zero.

This ratio amplifies the anomaly signal for regions both dissimilar from normal patterns (high $s_N$) and similar to known anomaly patterns (low $s_P$).

\subsubsection{Anomaly Localization}

For pixel-level anomaly segmentation, we extend our dual memory bank approach to generate spatial anomaly maps:
\begin{align}
S_N(h,w) &= \min_{m \in \mathcal{M}_N} \|\phi_{\text{test},\{2,3\}}(\mathcal{N}_p^{(h,w)}) - m\|_2 \\
S_P(h,w) &= \min_{m \in \mathcal{M}_P} \|\phi_{\text{test},\{2,3\}}(\mathcal{N}_p^{(h,w)}) - m\|_2
\end{align}

After applying neighborhood-aware weighting to obtain $S_N^w(h,w)$ and $S_P^w(h,w)$, we fuse these maps:
\begin{equation}
S_{\text{ratio}}(h,w) = \frac{S_N^w(h,w)}{S_P^w(h,w) + \epsilon}
\end{equation}

The resulting map is upsampled to match the original image dimensions and smoothed with a Gaussian filter ($\sigma = 2$) to enhance visual clarity while preserving fine details.

\section{Experiments}
\label{sec:experiment}
\subsection{Dataset}
We evaluate our method on the KSDD2 dataset~\cite{bovzivc2021mixed}, a real-world industrial benchmark for surface defect detection. The dataset contains 2,085 normal and 246 defective training images, along with 894 normal and 110 defective test images. Defects include scratches, spots, and material imperfections, ranging from 0.5 cm to 15 cm, captured under factory conditions. All images are resized to 224 × 632 pixels to standardize resolution while preserving defect morphology and spatial context.

\subsection{Implementation Details}
Experiments used an NVIDIA RTX 4090 GPU with a WideResNet50 backbone (ImageNet pretrained). We implemented 3×3 patches, feature hierarchies from ResNet levels 2-3, coreset subsampling (1\% negative, 10\% positive memory banks), and k=3 neighborhood weighting. Synthetic anomalies were generated using SDXL~\cite{podell2023sdxl} via Diffusers~\cite{von2022diffusers}, with text prompts "white marks on the wall" and "copper metal scratches" alongside a negative prompt "smooth, plain, black, dark, shadow" to suppress artifacts. Using KSDD2 ground-truth masks, defect-free training images were inpainted to create context-preserving synthetic anomalies, which were combined with the original training set. All models were implemented in PyTorch.

\section{Results}

We evaluated our \methname framework on the KSDD2 dataset using both image-level and pixel-level metrics to comprehensively assess detection and localization capabilities. Table~\ref{tab:performance_comparison} presents a comparative analysis with state-of-the-art methods in industrial anomaly detection.

Our experiments reveal that \methname outperforms most existing methods across both detection and localization tasks. While IRP achieves comparable image-level detection performance (94.0\% vs. our 94.2\%), it does not provide pixel-wise localization capabilities, which are crucial for practical industrial applications. Both OSR and IRP lack localization ability entirely, as indicated by the missing pixel-wise AUROC values.
The \methname base configuration already surpasses PatchCore by 1.9\% in image-level AUROC and 1.1\% in pixel-wise AUROC, demonstrating the effectiveness of our dual memory bank architecture. When implemented with the full configuration including synthetic data augmentation, \methname achieves state-of-the-art performance across both metrics. The 1.1\% improvement from base to full configuration highlights the value of our synthetic data approach in enhancing both detection and localization capabilities.
Notably, our method substantially outperforms earlier approaches like DRAEM and DSR, which struggle particularly with pixel-wise localization (42.4\% and 61.4\% respectively, compared to our 97.7\%).
\vspace{-10pt}
\begin{table}[H]
\centering
\caption{Anomaly Detection and Localization Performance on KSDD2 Dataset. \methname (base) denotes our dual memory bank architecture without synthetic data, while \methname (full) includes diffusion-based synthetic augmentation.}
\label{tab:performance_comparison}
\adjustbox{max width=\textwidth}{
\begin{tabular}{lccccccc}
\hline
\textbf{Metric} & \textbf{DRÆM~\cite{zavrtanik2021draem}} & \textbf{DSR~\cite{zavrtanik2022dsr}} & \textbf{PatchCore~\cite{roth2022towards}} & \textbf{OSR~\cite{aqeel2024delta}} & \textbf{IRP~\cite{Aqeel_2025}} & \textbf{\methname (base)} & \textbf{\methname (full)} \\ \hline
\textbf{I-AUROC (\%)} & 77.8 & 87.2 & 91.2 & 92.1 & 94.0 & 93.1 & 94.2 \\ \hline
\textbf{P-AUROC (\%)} & 42.4 & 61.4 & 95.8 & - & - & 96.9 & 97.7 \\ \hline
\end{tabular}
}
\end{table}

\vspace{-3em}
\subsection{Augmentation Analysis}
To isolate and quantify the effect of our synthetic data augmentation strategy within the \methname framework, we conducted experiments with varying numbers of synthetic samples, as shown in Table~\ref{tab:augmented_samples}. With no synthetic samples, the \methname base configuration relies solely on the limited real defective samples (246) available in the KSDD2 dataset alongside 2,085 normal samples. As synthetic samples are introduced, both detection and localization performance steadily improve. The addition of 100 synthetic samples (50 per text prompt) yields optimal performance across metrics. Interestingly, increasing the synthetic sample count to 150 provides no additional benefits and slightly reduces performance, suggesting a saturation point in the diversity of synthetic defect characteristics. These results validate our \methname approach of integrating synthetic models for industrial anomaly detection, demonstrating that carefully generated synthetic defects can effectively supplement limited real-world data while preserving the industrial context necessary for accurate detection and localization.

\vspace{-1em}
\begin{table}[H]
\centering
\caption{Effect of varying the number of augmented samples on \methname performance.}
\label{tab:augmented_samples}
\adjustbox{max width=\textwidth}{
\begin{tabular}{lcccc}
\hline
\textbf{Augmented Images} & \textbf{0} & \textbf{50} & \textbf{100} & \textbf{150} \\
\hline
\textbf{I-AUROC (\%)} & 93.1 & 93.4 & 94.2 & 93.5 \\
\hline
\textbf{P-AUROC (\%)} & 96.9 & 97.1 & 97.7 & 97.2 \\
\hline
\end{tabular}
}
\end{table}

\vspace{-3em}
\subsection{Qualitative Analysis}
\begin{figure}[tb]
  \centering
  \includegraphics[width=0.8\textwidth]{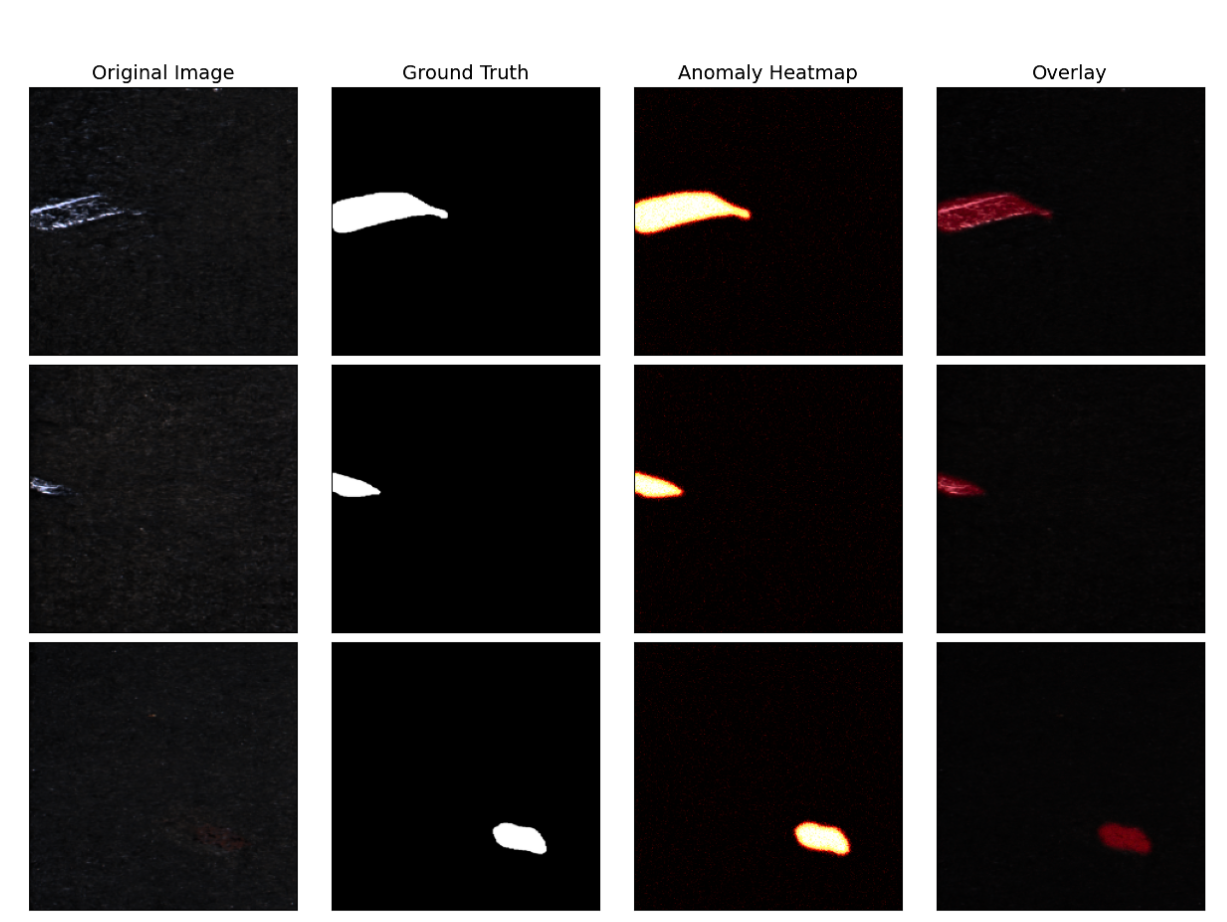}
  \caption{Qualitative comparison of anomaly localization results on the KSDD2 test set.}
  \label{fig:augment}
\end{figure}

The visualization in Figure~\ref{fig:augment} demonstrates that \methname effectively detects and localizes various types of defects in electrical commutators, including subtle scratches, surface anomalies, and material imperfections. When comparing the heatmaps generated by the standard PatchCore approach with those from the \methname base and full configurations, we observe significantly better alignment with ground truth masks and a reduction in false positives in background regions. The full \methname implementation produces anomaly maps with sharper boundaries and improved detection of subtle defect patterns, which aligns with the quantitative improvements observed in our experimental results.

\vspace{-10pt}

\section{Conclusion}
\methname represents a significant advancement in industrial anomaly detection by reconceptualizing defects as occupying structured feature distributions rather than arbitrary deviations. The integration of explicit dual-distribution modeling with diffusion-based synthetic defect generation creates a robust framework that leverages limited anomaly data effectively. The empirical performance ceiling observed at 100 synthetic samples suggests an optimal balance between augmentation diversity and potential distribution shift. This work establishes a foundation for future research in adaptive memory dynamics and uncertainty quantification for defect detection in data-constrained industrial environments, particularly for applications requiring precise boundary delineation and reduced false positives.

\section*{Acknowledgements}
This study was carried out within the PNRR research activities of the consortium iNEST (Interconnected North-Est Innovation Ecosystem) funded by the European Union Next-GenerationEU (Piano Nazionale di Ripresa e Resilienza (PNRR) – Missione 4 Componente 2, Investimento 1.5 – D.D. 1058  23/06/2022, ECS\_00000043).

\par\vfill\par

%
%
\bibliographystyle{splncs04}
\bibliography{main}
\end{document}